\documentclass{article}
\usepackage{newstyle}
\usepackage{amsmath,graphicx,amssymb,algorithm,algorithmic}
\usepackage{color}

\newcommand{\fig}[1]{Fig. (\ref{#1})}

\title{NeuNetS: An Automated Synthesis Engine for Neural Network Design}

\name{
\begin{tabular}{c}
Atin Sood,$^{\star,1}$ 
Benjamin Elder,$^{\star,2}$ 
Benjamin Herta,$^{\dagger,4}$
Chao Xue,$^{\dagger,6}$ 
Costas Bekas,$^{\dagger,3}$
\protect\\
A.\ Cristiano I.\ Malossi,$^{\dagger,3}$ 
Debashish Saha,$^{\dagger,4}$
Florian Scheidegger,$^{\dagger,3}$
Ganesh Venkataraman,$^{\dagger,4}$ 
\protect\\
Gegi Thomas,$^{\dagger,4}$
Giovanni Mariani,$^{\dagger,3}$
Hendrik Strobelt,$^{\dagger,2}$
Horst Samulowitz,$^{\dagger,4}$
\protect\\
Martin Wistuba,$^{\dagger,5}$
Matteo Manica,$^{\dagger,3}$
Mihir Choudhury,$^{\dagger,4}$
Rong Yan,$^{\dagger,6}$ 
Roxana Istrate,$^{\dagger,3}$ 
\protect\\
Ruchir Puri,$^{\star,4}$
Tejaswini Pedapati\,$^{\dagger,4}$ 
\end{tabular},
}

\address{
\protect\large{$^\star$IBM Watson AI Platform \hspace{0.10cm} and \hspace{0.10cm} $^\dagger$IBM Research AI}
\vspace{0.2cm}
\\$^1$New York, NY, USA \hspace{0.5cm} $^2$ Cambridge, MA, USA \hspace{0.5cm}
$^3$R\"uschlikon, Zurich, Switzerland    \\$^4$Yorktown Heights, NY, USA \hspace{0.5cm} 
$^5$Dublin, Ireland \hspace{0.5cm}
$^6$Beijing, China
}

\begin{document}

\ninept

\maketitle

\begin{abstract}
Application of neural networks to a vast variety of practical applications is transforming the way AI is applied in practice. Pre-trained neural network models available through APIs or capability to custom train pre-built neural network architectures with customer data has made the consumption of AI by developers much simpler and resulted in broad adoption of these complex AI models. While pre-built network models exist for certain scenarios, to try and meet the constraints that are unique to each application, AI teams need to think about developing custom neural network architectures that can meet the tradeoff between accuracy and memory footprint to achieve the tight constraints of their unique use-cases. However, only a small proportion of data science teams have the skills and experience needed to create a neural network from scratch, and the demand far exceeds the supply. In this paper, we present NeuNetS : An automated Neural Network Synthesis engine for custom neural network design that is available as part of IBM's AI OpenScale's product. NeuNetS is available for both Text and Image domains and can build neural networks for specific tasks in a fraction of the time it takes today with human effort, and with accuracy similar to that of human-designed AI models.
\vspace{0.1cm}
\end{abstract}
\section{Introduction}
\label{sec:intro}
AI is changing the way businesses work. However, it’s important to remember that every business has unique challenges to solve, and the range of use-cases for AI is constantly expanding. While pre-built AI models exist for certain scenarios, to try and meet the constraints that are unique to each application, AI teams will need to think about developing custom AI models of their own. Artificial neural networks are arguably the most powerful tool currently available to data scientists and businesses. However, only a small proportion of data scientists have the skills and experience needed to create a neural network from scratch, and the demand far exceeds the supply. As a result, getting a new neural network that is architecturally custom designed to meet the needs of that application, even to the proof-of-concept stage, requires a level of investment that most enterprises struggle to afford. Automation technologies that bridge this skills gap by automatically designing the architecture of neural networks for a given data are increasingly gaining importance. In this paper, we present NeuNetS : A Neural Network Synthesis engine for neural network design that is available as part of IBM's AI OpenScale's product. NeuNetS automatically configures itself to the needs of the user and the use case and helps reduce the complexity and skills required to build AI models, making data science teams more productive and enabling them to scale AI across their workflows. Overall, NeuNetS has two main stages: Coarse-grained synthesis and Fine-grained synthesis. Coarse-grained synthesis automatically optimizes and determines the overall architecture of the network: How many layers there should be, how they are connected, different architectural features like convolution layers and so on. The unique and novel step of fine-grained synthesis enables NeuNetS to take a deeper dive into each layer and optimizes the individual neurons and connection—for example, what kind of convolution filter should be applied, and which neurons and edges should be optimized. One of the critical breakthroughs that have enabled this capability is a very high-fidelity approach to performance estimation, which allows us to bypass real-time training and analysis and design neural networks automatically in a matter of hours—compared to the weeks or months that it might take a data scientist to train and optimize the AI model. NeuNetS is available for both Text and Image domains and can build neural networks for specific tasks in a fraction of the time it takes today, and with accuracy similar to that of human-designed AI models. The data science teams can then further fine tune the model, leading to greater productivity and cost-efficiency. NeuNetS is a novel tool for augmenting human expertise with powerful, AI-driven optimization capabilities.

The remainder of the paper is organized as follows.  We first provide the underlying flexible architecture of NeuNetS in Sec.~\ref{sec:arch}.  Next, in Sec.~\ref{sec:coarse-grained-algo}, we give details behind two of the coarse grained neural architecture search engines that are key part of NeuNetS. In Sec.~\ref{sec:fine-grained-algo}, we describe a unique set of fine-grained transformation to further optimize the Neural Network designs.  Finally, in Sec.~\ref{sec:experiments}, we provide empirical results on several standard and real-world datasets.  

\section{Related Work}

Evolutionary algorithms and reinforcement learning are currently the two state-of-the-art techniques used by neural network architectures search algorithms.
With Neural Architecture Search \cite{DBLP:journals/corr/ZophL16}, Zoph et al. demonstrated in an experiment over 28 days and with 800 GPUs that neural network architectures with performances close to state-of-the-art architectures can be found.
In parallel or inspired by this work, others proposed to use reinforcement learning to detect sequential architectures \cite{baker}, reduce the search space to repeating cells \cite{zoph2017learning,DBLP:journals/corr/abs-1708-05552} or apply function-preserving actions to accelerate the search \cite{cai2018efficient}.

Neuro-evolution dates back three decades.
In the beginning it focused only on evolving weights \cite{Miller1989} but it turned out to be effective to evolve the architecture as well \cite{Stanley2002}.
Neuro-evolutionary algorithms gained new momentum due to the work by Real et al. \cite{real2017large}.
In an extraordinary experiment that used 250 GPUs for almost 11 days, they showed that architectures can be found which provide similar good results as human-crafted image classification network architectures.
Very recently, the idea of learning cells instead of the full network has also been adopted for evolutionary algorithms \cite{Liu18b}.
Miikkulainen et al. even propose to coevolve a set of cells and their wiring \cite{DBLP:journals/corr/MiikkulainenLMR17}.

Other methods that try to optimize neural network architectures or their hyperparameters are based on model-based optimization \cite{Snoek2012,Liu2018a,Diaz2017,Wistuba2017a} and Monte-Carlo Tree Search \cite{negrinho2017deeparchitect, Wistuba2017, Wang2018}.

Various techniques exist which try to shorten the training time.
One idea is based on the idea of terminating unpromising training runs early.
The partially observed learning curve is used directly to decide to terminate a run early \cite{Li17} or first extrapolated and then used~\cite{lce,bnn,svr}.
Other methods are able to sample different architectures and then predict its likely performance.
Peephole~\cite{peephole} predicts a network accuracy by only analyzing the network structure, however it works only on a fixed dataset test case.
SMASH uses a hypernetwork to predict weights for an architecture without training and uses its validation performance as a proxy for its performance after training~\cite{Brock2017}.
Others reduce the search time by sharing or reusing model weights~\cite{cai2018efficient,Pham2018,Liu18,Wistuba2018}.

\section{NeuNetS Architecture}\label{sec:arch}

\subsection{Overview}\label{sec:arch-overview}

\begin{figure}
    \centering
    \includegraphics[scale=0.3]{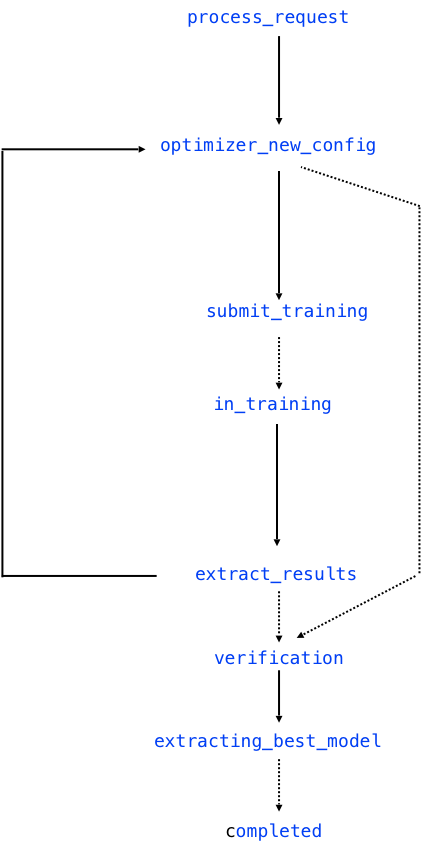}
    \caption{Execution Pipeline Operational States}
    \label{fig:execution-pipeline-operational-states}
\end{figure}

The lifecycle of a NeuNetS project consists of a series of states or stages, as detailed in \fig{fig:execution-pipeline-operational-states}. During this lifecycle, the synthesis states are executed multiple times to explore/evolve, train, and evaluate different networks. Once stopping conditions, whether budgetary or algorithmic, are reached, the synthesis loop ends and final results are extracted to the user's storage instance.

\begin{figure}
    \centering
    \includegraphics[scale=0.1]{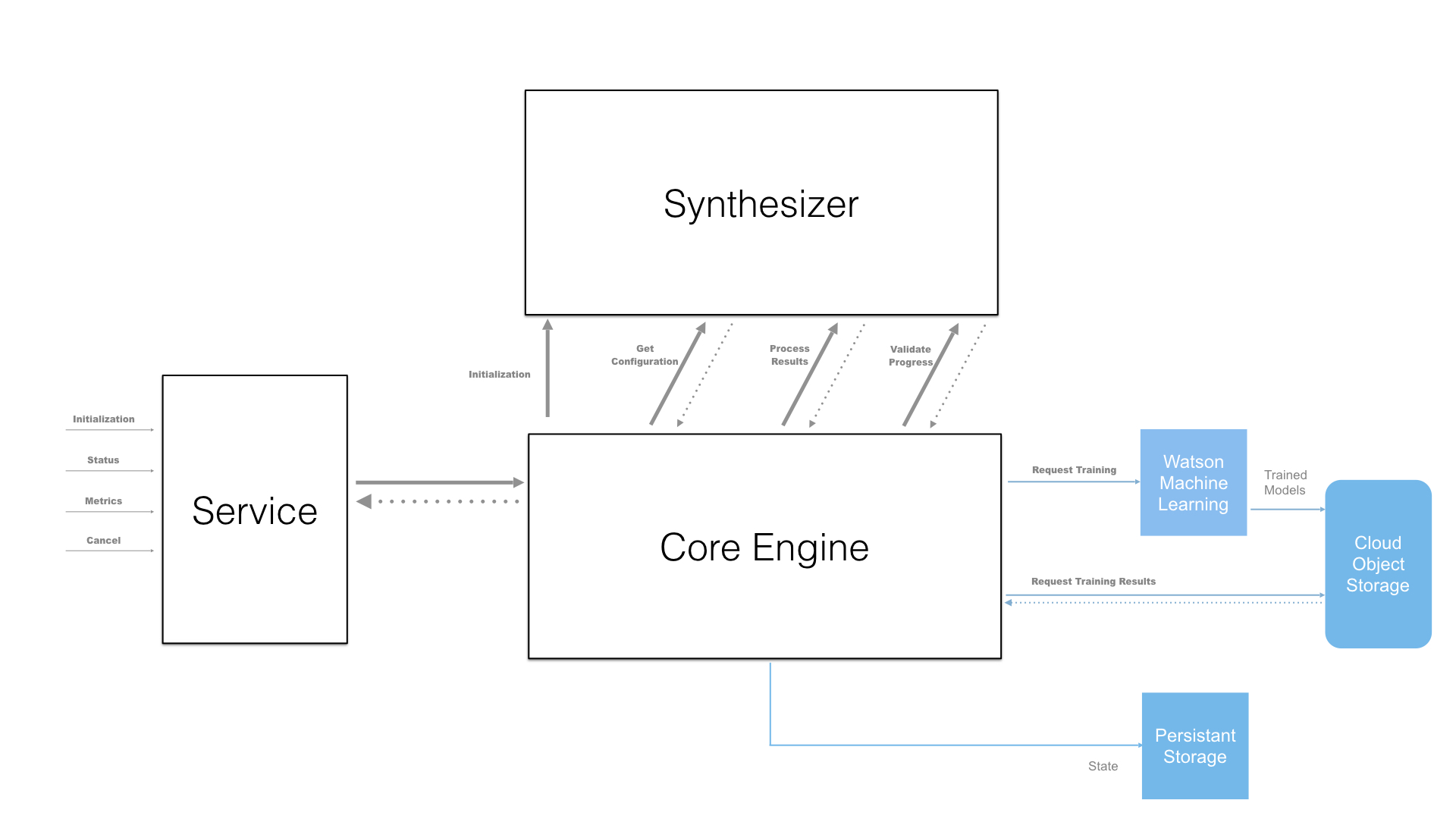}
    \caption{NeuNetS Component Architecture}
    \label{fig:neunets-component-architecture}
\end{figure}

The architectural implementation of NeuNetS consists of three main components: the service component, the core engine component, and the synthesizer component. The relations between these components and the required external services is illustrated in \fig{fig:neunets-component-architecture}. The service component includes the NeuNetS APIs and handles all incoming requests to the NeuNetS project. The core engine component maintains the state of the project and other relevant data. In each synthesis cycle, it obtains new architecture configurations from the synthesizer component and submits them to Watson Machine Learning \cite{ibm-wml} for training. When the stopping conditions are reached, it stores the final models in the user's cloud storage instance. The synthesizer component is a pluggable register of algorithms which use the state information passed from the engine to produce new architecture configurations. The rest of this section describes the functionality of these components in more detail. 

\subsection{Service Component}
\label{sec:arch-services}

The NeuNetS service component receives and manages all API requests and responses. These include initialization of a NeuNetS operation, obtaining the ongoing status, providing metrics, and prematurely stopping an operation. An incoming request to initialize and start an operation results in a preliminary validation of the request parameters. The service component performs a series of internal calculations to determine optimal operation parameters with which to initialize the NeuNetS synthesis. Once these preliminary checks are passed, the service component calls the core engine to initialize a synthesis pipeline. 

\begin{figure}
    \centering
    \includegraphics[width=\linewidth]{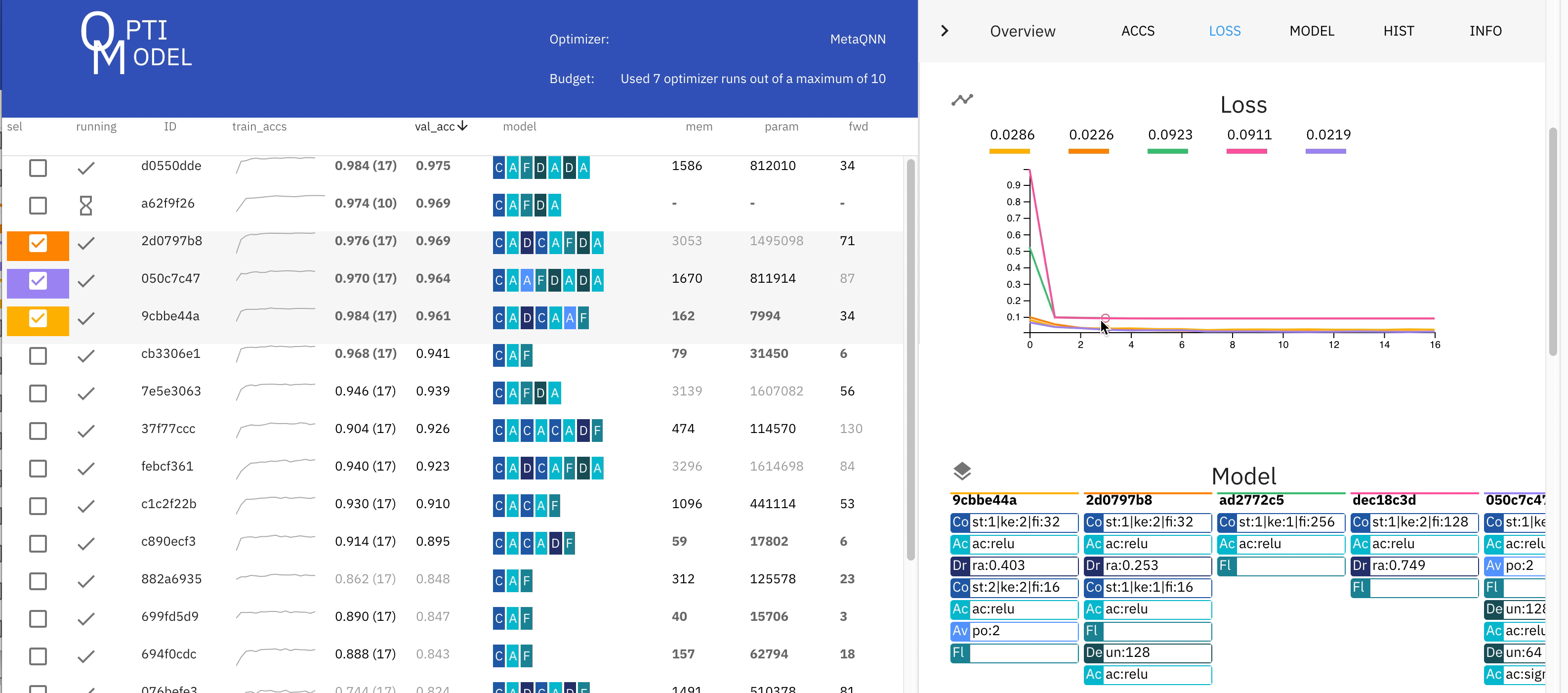}
    \caption{NeuNetS visual interface to manually pick a best model. Left: list of models in training and trained models with measures of performance and architecture diagram. Right: Details to compare selected models in depth. }
    \label{fig:optimodel}
\end{figure}

One of the unique attributes of NeuNetS (not available in the beta feature) is an API providing advanced visualization techniques for synthesized models. As illustrated in \fig{fig:optimodel}, these tools allow users to interactively compare various metrics of the models that they train with NeuNetS.

\subsection{Core Engine Component}
\label{sec:arch-core}

The NeuNetS core engine component is responsible for the overall lifecycle management of a NeuNetS operation. When it receives a request from the services component, it initializes a synthesis pipeline. The engine then calls the synthesizer component to obtain an initial set of architecture configurations. It processes these configurations and submits them to the Watson Machine Learning Service to be trained. This requires that the core engine define and compute all required resources (GPU/CPU, memory) for the training. Once the configuration has been trained for the requested amount time, the engine stores these intermediate configurations in an internal persistent storage instance. These configurations, along with their performance metrics from training, are then provided to the synthesizer component, which uses this information to produce a new set of configurations. This loop is performed multiple times to iteratively improve the network performance. At each step the engine checks the validity of the results from the last step, handles any error conditions, places relevant data into the persistent storage, and updates the state of the pipeline.

This process terminates either when the engine determines that specific budgetary objectives have been achieved, or the synthesis component receives a configuration whose performance meets its algorithmic requirements. A final completion operation releases all pipeline resources and facilitates the transfer of the final synthesized neural network model to a destination bucket inside of a valid IBM Cloud Object Storage \cite{ibm-cos} service instance.

The communication with persistent storage is a key operational aspect of the core engine. This decouples the overall state of the pipeline from the components and services performing the synthesis. This decoupling enables operational recovery in the event of service component failure, as new instances of the service component can immediately resume managing the lifecycle of the active pipeline, based on the stored pipeline state data.

\subsection{Synthesizer Component}
\label{sec:arch-optimizers}

The NeuNetS synthesizer component provides a pluggable framework for multiple distinct model synthesis algorithms, each of which get registered with the NeuNetS core engine component. Each algorithm implements a common base interface that reflects the required interaction between itself and the core engine. An initialization contract in the interface provides the algorithm with overall synthesis parameters specific to a provided dataset, as well as runtime parameters related to an executing pipeline’s environmental budgetary considerations. The interface also defines the contracts for three important operations that govern a NeuNetS pipeline operation. The first operation encompasses the algorithm providing a configuration to the NeuNetS core engine that represents one or more data-inspired architectures of a deep layer neural network. The second operation is centered on providing the algorithm with all training results, along with any associated artifacts of consequence. The third operation is a validation operation, whereby the NeuNetS core engine queries the algorithm for an overall assessment of the pipeline progress based on the training results and additional state available to the algorithm.

\section{Neural Network Synthesis Methods}

\subsection{Coarse grained synthesis}
\label{sec:coarse-grained-algo}
NeuNetS features three large scale architecture search algorithms: NCEvolve~\cite{Wistuba2018}, TAPAS~\cite{tapas2019}, and Hyperband++. These algorithms make a step forward with respect to the most advanced works in the literature, addressing fundamental problems such as dataset generality and performance scalability. 

NeuNetS algorithms are designed to synthesize new models in a short and reasonable time, without using transfer-learning or pre-trained models. This allows us to explore a wide space of network architecture configurations, and fine-tune the model for the specific dataset provided by the user. 

Being based on multiple optimization algorithms, NeuNetS can accommodate a wider range of model synthesis scenarios. In future releases, the user will not only be able to update data, but also to decide how much time and how many resources to allocate for the model synthesis, as well as optionally the maximum size of the model, and the target deployment platform. Based on these constraints, NeuNetS will select the best optimization strategy to serve back to the users the right models for their needs.

The portfolio of algorithms will be continuously extended including top works from the public community, as well as further advanced developments from IBM Research.

In the following paragraphs, we briefly recall the main technical features of our current portfolio of optimization algorithms.

\subsubsection{NCEvolve}
NCEvolve is a novel neuro-evolutionary technique to search for neural architectures without human interference.
It assumes that a neural network architecture is a sequence of neuro-cells and keeps mutating them using function-preserving operations.
This assumption has several advantages.
First, it reduces the search space complexity.
Second, these cells are possibly transferable and can be used in order to arbitrarily extend the complexity of the network.
Mutations based on function-preserving operations guarantee better parameter initialization than random initialization such that less training time is required per network architecture.

Chen et al. \cite{Chen2016} proposed a family of function-preserving network manipulations in order to transfer knowledge from one network to another.
Suppose a teacher network is represented by a function $f\left(\mathbf{x}\ |\ \boldsymbol{\theta}^{\left(f\right)}\right)$ where $\mathbf{x}$ is the input of the network and $\boldsymbol{\theta}^{\left(f\right)}$ are its parameters.
Then an operation changing the network $f$ to a student network $g$ is called function-preserving if and only if the output for any given model remains unchanged:
\begin{equation}
\forall\mathbf{x}:\ f\left(\mathbf{x}\ |\ \boldsymbol{\theta}^{\left(f\right)}\right) =g\left(\mathbf{x}\ |\ \boldsymbol{\theta}^{\left(g\right)}\right)\enspace.
\end{equation}
Note that typically the number of parameters of $f$ and $g$ are different.
We will use this approach in order to initialize our mutated network architectures.
Then, the network is trained for some additional epochs with gradient-based optimization techniques.
Using this initialization, the network requires only few epochs before it provides decent predictions.
We briefly explain the proposed manipulations and our novel contributions to it.
Please note that a fully connected layer is a special case of a convolutional layer.
For a more detailed description, we refer to \cite{Wistuba2018}.

\subparagraph{Convolutions in Deep Learning}
Convolutional layers are a common layer type used in neural networks for visual tasks.
We denote the convolution operation between the layer input $X\in\mathbb{R}^{w\times h\times i}$ with a layer  with parameters $W\in\mathbb{R}^{k_{1}\times k_{2}\times i\times o}$ by $X\ast W$.
Here, $i$ is the number of input channels, $w\times h$ the input dimension, $k_{1}\times k_{2}$ the kernel size and $o$ the number of output feature maps.
Depthwise separable convolutions, or for short just separable convolutions, are a special kind of convolution factored into two operations.
During the depthwise convolution a spatial convolution with parameters $W_{d}\in\mathbb{R}^{k_{1}\times k_{2}\times i}$ is applied for each channel separately.
We denote this operation by using $\circledast$.
This is in contrast to the typical convolution which is applied across all channels.
In the next step the pointwise convolution, i.e.
a convolution with a $1\times1$ kernel, traverses the feature maps which result from the first operation with parameters $W_{p}\in\mathbb{R}^{1\times1\times i\times o}$.
Comparing the normal convolution operation $X\ast W$ with the separable convolution $\left(X\circledast W_{d}\right)\ast W_{p}$, we immediately notice that in practice the former requires with $k_{1}k_{2}io$ more parameters than the latter which only needs $k_{1}k_{2}i+io$.

\subparagraph{Layer Widening}

Assume the teacher network $f$ contains a convolutional layer with a $k_{1}\times k_{2}$ kernel which is represented by a matrix $W^{\left(l\right)}\in\mathbb{R}^{k_{1}\times k_{2}\times i\times o}$ where $i$ is the number of input feature maps and $o$ is the number of output feature maps or filters.
Widening this layer means that we increase the number of filters to $o'>o$.
Chen et al. \cite{Chen2016} proposed to extend $W^{\left(l\right)}$ by replicating the parameters along the last axis at random.
This means the widened layer of the student network uses the parameters
\begin{equation}
V_{\cdot,\cdot,\cdot,j}^{\left(l\right)}=
\begin{cases}
W_{\cdot,\cdot,\cdot,j}^{\left(l\right)} & j\leq o\\ W_{\cdot,\cdot,\cdot,r}^{\left(l\right)} & r\text{ uniformly sampled from }\left\{ 1,\ldots,o\right\}
\end{cases}.\label{eq:widening_student_w1}
\end{equation}
In order to achieve the function-preserving property, the replication of some filters needs to be taken into account for the next layer $V^{\left(l+1\right)}$.
This is achieved by dividing the parameters of $W_{\cdot,\cdot,j,\cdot}^{\left(l+1\right)}$ by the number of times the $j$-th filter has been replicated.
If $n_{j}$ is the number of times the $j$-th filter was replicated, the weights of the next layer for the student network are defined by
\begin{equation}
V_{\cdot,\cdot,j,\cdot}^{\left(l+1\right)}=\frac{1}{n_{j}}W_{\cdot,\cdot,j,\cdot}^{\left(l+1\right)}\enspace.\label{eq:widening_student_w2}
\end{equation}
We extended this mechanism to depthwise separable convolutional layers.
A depthwise separable convolutional layer at depth $l$ is widened as follows.
The pointwise convolution for the student is estimated according to Equation \ref{eq:widening_student_w1}.
This results into replicated output feature maps.
The depthwise convolution is identical to the one of the teacher network, i.e. the operations with parameters $a$ and $b$.
Independently of whether we used a depthwise separable or normal convolution in layer $l$, widening it requires adaptations in a following depthwise separable convolutional layer.
The parameters of the depthwise convolution are replicated according to the replication of parameters in the previous layer similar to Equation \ref{eq:widening_student_w1}.
Furthermore, the parameter of the pointwise convolution depend on the replications in the previous layers analogously to Equation \ref{eq:widening_student_w2}.

\subparagraph{Layer Deepening\label{subsec:Layer-Deepening}}

Chen et al. \cite{Chen2016} proposed a way to deepen a network by inserting an additional convolutional or fully connected layer.
We complete this definition by extending it to depthwise separable convolutions.

A layer can be considered to be a function which gets as an input the output of the previous layer and provides the input for the next layer.
A simple function-preserving operation is to set the weights of a new layer such that the input of the layer is equal to its output.
If we assume $i$ incoming channels and an odd kernel height and weight for the new convolutional layer, we achieve this by setting the weights of the layer with a $k_{1}\times k_{2}$ kernel to the identity matrix:
\begin{equation}
V_{j,h}^{\left(l\right)}=\begin{cases}
I_{i,i} & j=\frac{k_{1}+1}{2}\wedge h=\frac{k_{2}+1}{2}\\
\mathbf{0} & \text{otherwise}
\end{cases}\enspace.\label{eq:deepening_student_w1}
\end{equation}
This operation is function-preserving and the number of filters is equal to the number of input channels.
More filters can be added by layer widening, however, it is not possible to use less than $i$ filters for the new layer.
Another restriction is that this operation is only possible for activation functions $\sigma$ with
\begin{equation}
\sigma\left(\mathbf{x}\right)=\sigma\left(I\sigma\left(\mathbf{x}\right)\right)\ \forall\mathbf{x}\ .\label{eq:deepening_activation_function_definition}
\end{equation}
The ReLU activation function $\text{ReLU}\left(\mathbf{x}\right)=\max\left\{ \mathbf{x},\mathbf{0}\right\}$ fulfills this requirement.

We extend this operation to depthwise convolutions.
The parameters of the pointwise convolution $V_{p}$ are initialized analogously to Equation \ref{eq:deepening_student_w1} and the depthwise convolution $V_{d}$ is set to one: 
\begin{align}
V_{p} & =I_{i,i}\\
V_{d} & =\mathbf{1}\enspace.
\end{align}
This initialization ensures that both, the depthwise and pointwise convolution, just copy the input.
New layers can be inserted at arbitrary positions with one exception.
Under certain conditions an insertion right after the input layer is not function-preserving.
For example if a ReLU activation is used, there exists no identity function for inputs with negative entries.

\subparagraph{Kernel Widening}

Increasing the kernel size in a convolutional layer is achieved by padding the tensor using zeros until it matches the desired size.
The same idea can be applied to increase the kernel size of depthwise separable convolution by padding the depthwise convolution with zeros.

\subparagraph{Insert Skip Connections}

Many modern neural network architectures rely on skip connections \cite{He2016}.
The idea is to add the output of the current layer to the output of a previous.
One simple example is
\begin{equation}
X^{\left(l+1\right)}=\sigma\left(X^{\left(l\right)}\ast V^{\left(l+1\right)}+X^{\left(l\right)}\right)\enspace.\label{eq:skip-example}
\end{equation}
Therefore, we propose a function-preserving operation which allows inserting skip connection.
We propose to add layer(s) and initialize them in a way such that the output is 0 independent on the input.
This allows to add a skip because now adding the output of the previous layer to zero is an identity operation.
A new operation is added setting its parameters to zero, $V^{\left(l+1\right)}=\mathbf{0}$, achieving a zero output.
Now, adding this output to the input is an identity operation.

\subparagraph{Branch Layers\label{subsec:Branch-Layers}}

\begin{figure}[t]
\centering
\includegraphics[width=0.45\textwidth]{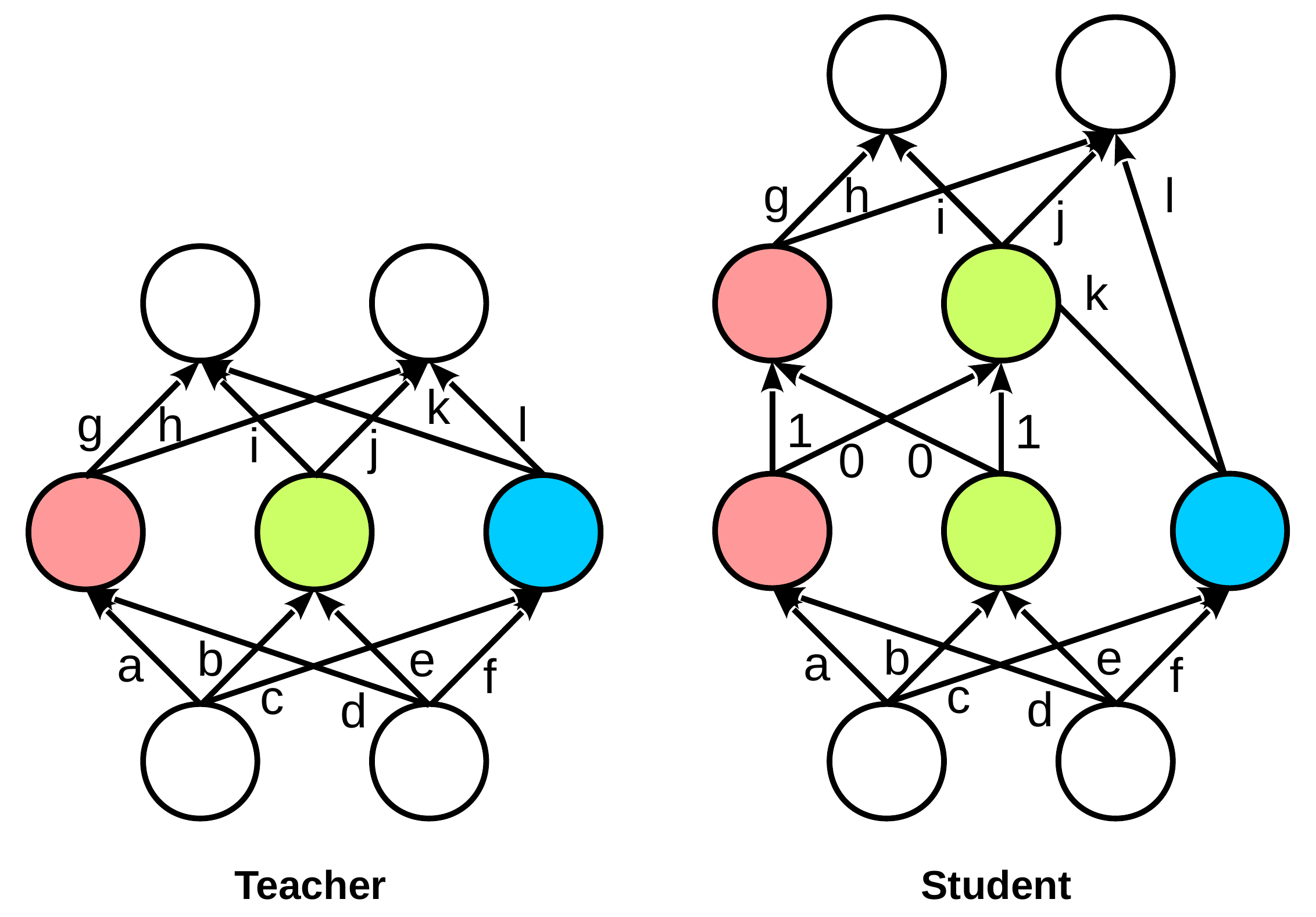}
\caption{
Visualization of branching the colored layer and insert a convolution into the left branch.
Same colored circles represent identical feature maps.
Circles without filling can have any value and are not important for the visualization.
Activation functions are omitted to avoid clutter.
\label{fig:net2net_branch_and_insert}}
\end{figure}

We also propose to branch layers.
Given a convolutional layer $X^{\left(l\right)}\ast W^{\left(l+1\right)}$ it can be reformulated as 
\begin{equation}
\text{merge}\left(X^{\left(l\right)}\ast V_{1}^{\left(l+1\right)},\ X^{\left(l\right)}\ast V_{2}^{\left(l+1\right)}\right)\enspace,
\end{equation}
where \emph{merge} concatenates the resulting output.
The student network's parameters are defined as
\begin{align*}
V_{1}^{\left(l+1\right)} & =W_{\cdot,\cdot,\cdot,1:\left\lfloor o/2\right\rfloor }^{\left(l+1\right)}\\
V_{2}^{\left(l+1\right)} & =W_{\cdot,\cdot,\cdot,\left(\left\lfloor o/2\right\rfloor +1\right):o}^{\left(l+1\right)}\enspace.
\end{align*}
This operation is not only function-preserving, it also does not add any further parameters and in fact is the very same operation.
However, combining this operation with other function-preserving operations allows to extend networks by having parallel convolutional operations or add new convolutional layers with smaller filter sizes.
In Figure \ref{fig:net2net_branch_and_insert} we demonstrate how to achieve this.
The colored layer is first branched and then a new convolutional layer is added to the left branch.
In contrast to only adding a new layer as described in Section~\ref{subsec:Layer-Deepening}, the new layer has only two output channels instead of three.

\subparagraph{Multiple In- or Outputs}

All the presented operations are still possible for networks where a layer might have inputs from different layers or provide output for multiple outputs.
In that case only the affected weights need to be adapted according to the aforementioned equations.

\paragraph{Evolution of Neuro-Cells}

\begin{figure}[t]
\centering
\includegraphics[width=0.45\textwidth]{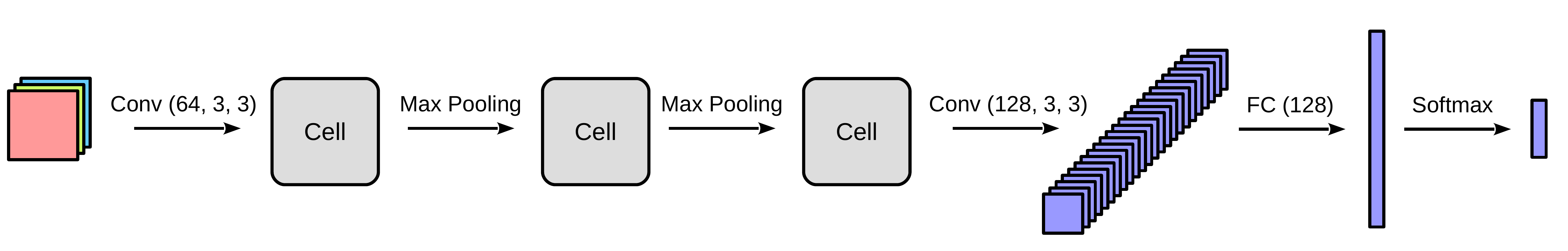}
\caption{Neural network template as used in our experiments.\label{fig:Neural-network-template}}
\end{figure}

The very basic idea of our proposed cell-based neuro-evolution is the following.
Given is a very simple neural network architecture which contains multiple neuro-cells (see Figure \ref{fig:Neural-network-template}).
The cells itself share their structure and the task is to find a structure that improves the overall neural network architecture for a given data set and machine learning task.
In the beginning, a cell is identical to a convolutional layer and is changed during the evolutionary optimization process.
Our evolutionary algorithm is using tournament selection to select an individual from the population:
randomly, a fraction $k$ of individuals is selected from the population.
From this set the individual with highest fitness is selected for mutation.
We define the fitness by the accuracy achieved by the individual on a hold-out data set.
The mutation is selected at random which is applied to all neuro-cells such that they remain identical.
The network is trained for some epochs on the training set and is then added to the population.
Finally, the process starts all over again.
After meeting some stopping criterion, the individual with highest fitness is returned.

\subparagraph{Mutations}

All mutations used are based on the function-preserving operations introduced in the last section.
This means, a mutation does not change the fitness of an individual, however, it will increase its complexity.
The advantage over creating the same network structure with randomly initialized weights is obviously that we start with a partially pretrained network.
This enables us to train the network in less epochs.
All mutations are applied only to the structure within a neuro-cell if not otherwise mentioned.
Our neuro-evolutional algorithm considers the following mutations.

\textit{Insert Convolution}
A convolution is added at a random position.
Its kernel size is $3\times3$, the number of filters is equal to its input dimension.
It is randomly decided whether it is a separable convolution instead.

\textit{Branch and Insert Convolution}
A convolution is selected at random and branched according to Section \ref{subsec:Branch-Layers}.
A new convolution is added according to the ``Insert Convolution'' mutation in one of the branches.
For an example see Figure \ref{fig:net2net_branch_and_insert}.

\textit{Insert Skip}
A convolution is selected at random.
Its output is added to the output of a newly added convolution (see ``Insert Convolution'') and is the input for the following layers.

\textit{Alter Number of Filters}
A convolution is selected at random and widened by a factor uniformly at random sampled from $\left[1.2,2\right]$.
This mutation might also be applied to convolutions outside of a neuro-cell.

\textit{Alter Number of Units}
Similar to the previous one but alters the number of units of fully connected layers.
This mutation is only applied outside the neuro-cells.

\textit{Alter Kernel Size}
Selects a convolution at random and increases its kernel size by two along each axis.

The motivation of selecting this set of mutations is to enable the neuro-evolutionary algorithm to discover similar architectures as proposed by human experts.
Adding convolutions allows to reach popular architectures such as VGG16 \cite{Simonyan2014}, combinations of adding skips and convolutions allow to discover residual networks \cite{He2016}.
Finally the combination of branching, change of kernel sizes and addition of (separable) convolutions allows to discover architectures similar to Inception \cite{Szegedy2015}, Xception \cite{Chollet2016} or FractalNet \cite{Larsson2017}.

The optimization is started with only a single individual.
Then always two individuals are selected with replacement based on the previously described tournament selection process and trained in parallel.

\subsubsection{TAPAS}

\begin{figure}[!t]
\includegraphics[width=0.5\textwidth]{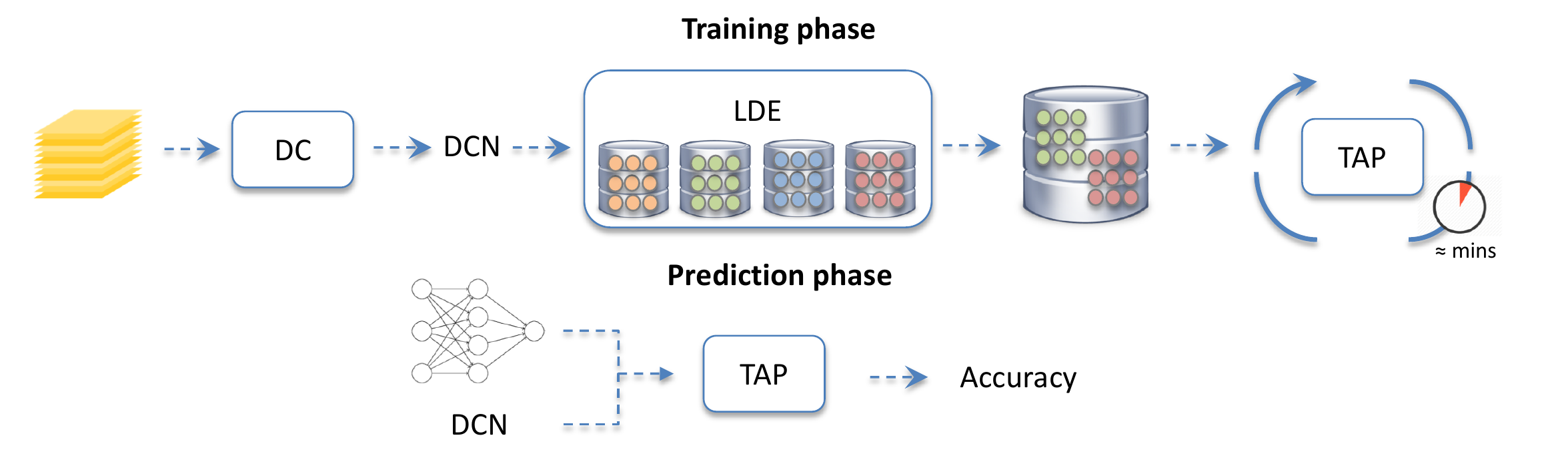}
\caption{Schematic TAPAS workflow. First row: the Dataset Characterization (DC) takes a new, unseen dataset and characterizes its difficulty by computing the Dataset Characterization Number (DCN). This number is then used to select a subset of experiments executed on similarly difficult datasets from the Lifelong Database of Experiments (LDE). Subsequently, the filtered experiments are used to train the Train-less Accuracy Predictor (TAP), an operation that takes up to a few minutes. Second row: the trained TAP takes the network architecture structure and the dataset DCN and predict the peak accuracy reachable after training. This phase scales very efficiently in a few seconds over a large number of networks.}
\label{fig:tapas_workflow}
\end{figure}

TAPAS is a framework that runs large scale architecture searches of thousands of networks in a few minutes on CPU. We achieve this with a novel deep neural network accuracy predictor, that estimates in fractions of a second classification performance for unseen input datasets, without training. In contrast to previously proposed approaches, our prediction is not only calibrated on the topological network information, but also on the characterization of the dataset-difficulty which  allows us to re-tune the prediction without any training. 
The TAPAS framework, depicted in Figure~\ref{fig:tapas_workflow}, is built on three main components:

\begin{enumerate}
\item \textbf{Dataset Characterization (DC):}
Receives an unseen dataset and computes a scalar score, namely the Dataset Characterization Number (DCN)~\cite{dataset_char}, which is used to rank datasets;

\item \textbf{Lifelong Database of Experiments (LDE):} 
Ingests training experiments of NNs on a variety of image classification datasets executed inside the TAPAS framework;

\item \textbf{Train-less Accuracy Predictor (TAP):} Given an NN architecture and a DCN, it predicts the potentially reachable peak accuracy without training the network.
\end{enumerate}

In the following we will detail each of the main components.

\paragraph{Dataset characterization (DC)}

The same CNN can yield different results if trained on an easy dataset (e.g., MNIST~\cite{data_mnist}) or on a more challenging one (e.g., CIFAR-100~\cite{data_cifar10_100}), although the two datasets might share features such as number of classes, number of images, and resolution. Therefore, in order to reliably estimate a CNN performance on a dataset we argue that we must first analyze the dataset difficulty. We compute the DCN by training a \emph{probe net} to obtain a dataset difficulty estimation~\cite{dataset_char}. We use the DCN for filtering datasets from the LDE and directly as input score in the TAP training and prediction phases as described in Section~\ref{sec:ap}.

\textbf{DCN computation} \emph{Prob nets} are modest-sized neural networks designed to characterize the difficulty of an image classification dataset~\cite{dataset_char}. We compute the DCN as peak accuracy, ranged in $[0, 1]$, obtained by training the \textit{Deep normalized ProbeNet} on a specific dataset for ten epochs. The DCN calculation cost is low due the following reasons: (i)~\textit{Deep norm ProbeNet} is a modest-size network, (ii)~the characterization step is performed only once at the entry of the dataset in the framework (the LDE stores the DCN afterwards), (iii)~the DCN does not require an extremely accurate training, thus reducing the cost to a few epochs, and (iv)~large datasets can be subsampled both in terms of number of images and of pixels.

\paragraph{Lifelong database of experiments (LDE)}\label{sec:LDE}

LDE is a continuously growing DB, which ingests every new experiment effectuated inside the framework. An experiment includes the CNN architecture description, the training hyper-parameters, the employed dataset (with its DCN), as well as the achieved accuracy. 

\begin{figure}[!t]
\centering
\includegraphics[width=0.5\textwidth]{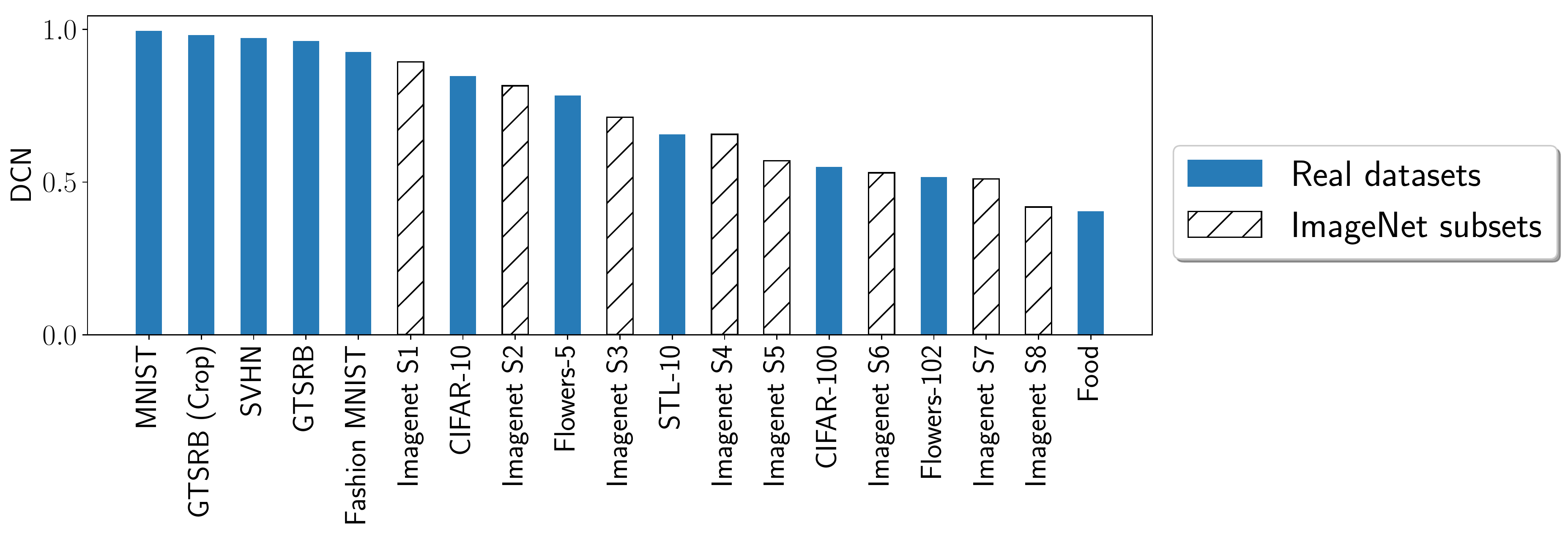}
\caption{List of image classification datasets used for characterization. The datasets are sorted by the DCN value from the easiest (left) to the hardest (right).
}
\label{fig:datasets}
\end{figure}

\textbf{LDE initialization} At the very beginning, the LDE is empty. Thus we perform a massive initialization procedure to populate it with experiments. 
For each available dataset in Figure~\ref{fig:datasets} we sample 800~networks from a slight variation of the space of MetaQNN~\cite{baker}. For convolution layers we use strides with values in $\{1, 2\}$, receptive fields with values in $\{3, 4,.. 256\}$, padding in $\{same, valid\}$ and whether is batch normalized or not. We also add two more layer types to the search space: residual blocks and skip connections. The hyperparameters of the residual blocks are the receptive field, stride and the repeat factor. The receptive field and the stride have the same bounds as in the convolution layer, while the repeat factor varies between 1 and 6 inclusively. The skip connection has only one hyperparameter, namely the previous layer to be connected to.

To speed up the process, we train the networks one layer at a time using the incremental method described in~\cite{incremental}. In this way we obtain the accuracies of all intermediary sub-networks at the same cost of the entire one. To facilitate the TAP, we train all networks with the same hyper-parameters, i.e., same optimizer, learning rate, batch size, and weights initiallizer. Although the fixed hyper-parameter setting seems a strong limitation and might limit peak accuracy by a few percent, it is enough to trim poorly performing networks and, in the case of an architecture search, to fairly rank competitive networks, the performance of which can later be  optimized further. As data augmentation we use standard horizontal flips, when possible, and left/right shifts with four pixels. For all datasets we perform feature-wise standardization.

This paper LDE initialization takes 18~months on a single P100~GPU. This number can be scaled down embarrassingly with the number of GPUs. It must also be considered that, even though the time spent to generate the LDE is comparable to the time of manual engineering search of hyperparameters, the LDE can then be employed in architecture searches for multiple datasets at no additional cost. Moreover, in an industrial environments, pre-existing runs on technical propriertary-datasets can be used to heat-up the LDE quickly.

\textbf{LDE selection} Let us consider an LDE populated with experiments from $N_d$~different datasets $D_j$, with $j=1,\dots,N_d$. Given a new input dataset~$\hat{D}$ and its corresponding characterization $\textrm{DCN}(\hat{D})$, the LDE block returns all experiments performed with datasets that satisfy the following relation
\begin{equation}
\| \textrm{DCN}(\hat{D}) - \textrm{DCN}(D_j) \| \leq \tau \qquad j \in [1,N_d],
\label{eq:florian_DCN}
\end{equation}
where $\tau$ is a predefined threshold that, in our experiments, is set to 0.05.

\paragraph{Train-less accuracy predictor (TAP)} \label{sec:ap}

\begin{figure}[!t]
\includegraphics[width=0.5\textwidth]{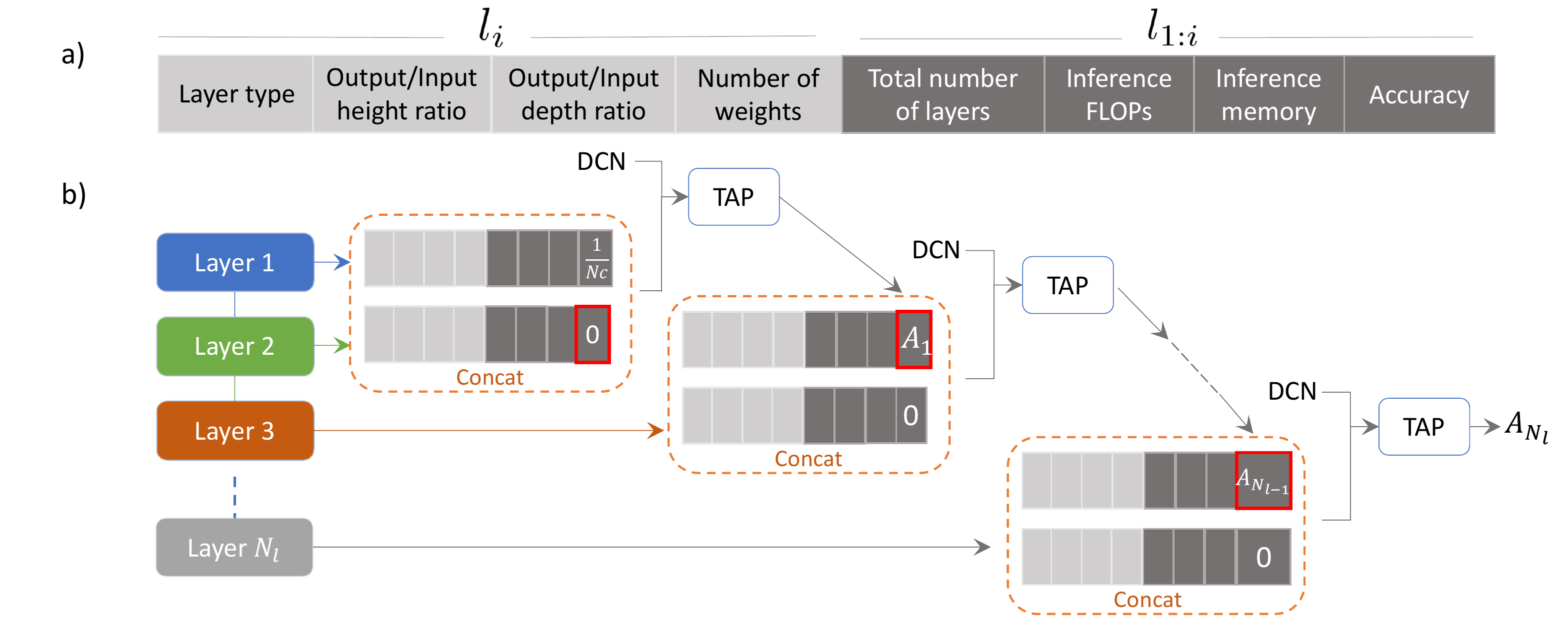}
\caption{Encoding vector structure and its usage in the iterative prediction. a) The encoding vector contains two blocks: $i$-layer information and from input to $i$-layer sub-network information. b) The encoding vector is used by the TAP following an iterative scheme. Starting from Layer~1 (input) we encode and concatenate two layers at a time and feed them to the TAP. In the concatenated vector, the \emph{Accuracy} field $\textrm{A}_i$ of $l_i$ is set to the predicted accuracy obtained from the previous TAP evaluation, whereas the one of $A_{i+1}$ corresponding to $l_{i+1}$ is always set to zero. For the input layer, we set $\textrm{A}_0$ to $1/N_c$, where $N_c$ is the number of classes, assuming a random distribution. The final predicted accuracy $A_{N_l}$ is the accuracy of the complete network.}
\label{fig:encoding}
\end{figure}

TAP is designed to perform fast and reliable CNN accuracy predictions. Compared to Peephole~\cite{peephole}, TAP leverages knowledge accumulated through experiments of datasets of similar difficulty filtered from the LDE based on the DCN. Additionally, TAP does not first analyze the entire NN structure and then makes a prediction, but instead performs an iterative prediction as depicted in Figure~\ref{fig:encoding}. In other words, it aims to predict the accuracy of a sub-network $l_{1:i+1}$, assuming the accuracy of the sub-network $l_{1:i}$ is known. The main building elements of the predictor are: (i) a compact encoding vector that represents the main network characteristics, (ii) a quickly-trainable network of LSTMs, and (iii) a layer-by-layer prediction mechanism. 

\textbf{Neural network architecture encoding} Similar to Peephole, TAP employs a layer-by-layer encoding vector as described in Figure~\ref{fig:encoding}. Unlike Peephole, we encode more complex information of the network architecture for a better prediction. 

Let us consider a network with $N_l$~layers, $l_i$ being the $i$-th layer counting from the input, with $i=1,\dots,N_l$. We define a CNN sub-network as $l_{a:b}$ with $1 \le a < b \le N_l$. Our encoding vector contains two types of information as depicted in Figure~\ref{fig:encoding}~a): (i)~$i$-th layer information and (ii)~$l_{1:i}$~sub-network information. For the current $i$-th layer we make the following selection of parameters: \emph{Layer type} is a one-hot encoding that identifies either convolution, pooling, batch normalization, dropout, residual block, skip connection, or fully connected. In future we will include latest motifs present in literature such as DenseNets \cite{huang2017densely} or AmoebaNets \cite{real2018regularized}. Note that for the shortcut connection of the residual block we use both the identity and the projection shortcuts~\cite{resnet}. The projection is employed only when the residual block decreases the number of filters as compared to the previous layer. Moreover, as compared to~\cite{peephole}, our networks do not follow a fixed skeleton in the convolutional pipeline, allowing for more generality. We only force a fixed block at the end, by using a global pooling and a fully connected layer to prevent networks from overfitting~\cite{nin}.

The \emph{ratio between the output height and input height} of each layer accounts for different strides or paddings, whereas the \emph{ratio between the output depth and input depth} accounts for modifications of the number of kernels. The \textit{number of weights} specifies the total of learnable parameters in $l_i$. This value helps the TAP differentiate between layers that increase the learning power of the network (e.g., convolution, fully connected layers) and layers that reduce the dimensionality or avoid overfitting (e.g., pooling, dropout). In the second part of the encoding vector, we include: \emph{Total number of layers}, counting from input to $l_i$, \emph{Inference FLOPs} and \emph{Inference memory} that are an accurate estimate of the computational cost and memory requirements of the sub-network, and finally \emph{Accuracy}, which is set either to 1/$N_c$, for the first layer, where $N_c$ is the number of classes to predict, zero for prediction purposes, or a specific value $\mathrm{A}_i \in [0, 1]$ that is obtained from the previous layer prediction. Before training, we perform a feature-wise standardization of the data, meaning that for each feature of the encoding vector, we subtract the mean and divide by the standard deviation.

\textbf{TAP architecture} TAP is a neural network consisting of two stacked LSTMs of 50 and 100 hidden units, respectively, followed by a single-output fully connected layer with sigmoid activation. The TAP network has two inputs. The first input is a concatenation of two encoding vectors corresponding to layer $l_i$ and $l_{i+1}$, respectively. This input is fed into the first LSTM. The second input is the DCN and is concatenated with the output of the second LSTM and then fed into the fully connected layer.

\textbf{TAP training} TAP requires a significant amount of training data to make reliable predictions. The LDE provides this data as described in Section~\ref{sec:LDE}. As mentioned above, all our generated networks are trained in an incremental fashion, as presented in \cite{incremental}, meaning that for each network of length $N_l$ we train all intermediary sub-networks $l_{1:k}$ with $1 < k \leq N_l$ and save their performance $A_{k}$. We encode each set of two consecutive layers $l_i$ and $l_{i+1}$ following the schema detailed in~\ref{sec:ap}, setting the accuracy field in the encoding vector of $l_i$ to $A_i$, which was obtained through training, and aiming to predict $A_{i+1}$.

TAP is trained with RMSprop~\cite{rmsprop}, using a learning rate of $10^{-3}$, a HeNormal weight initialization~\cite{heinit}, and a batch size of 512. As the architecture of the TAP is very small, the training process is of the order of a few minutes on a single GPU device. Moreover, the trained TAP can be stored and reapplied to other datasets with similar DCN numbers without the need for retraining.

\textbf{TAP prediction} TAP employs a layer-by-layer prediction mechanism. The accuracy~$\textrm{A}_{i}$ of the sub-network~$l_{1:i}$ predicted by the previous TAP evaluation is subsequently fed as input into the next TAP evaluation, which returns the predicted accuracy~$\textrm{A}_{i+1}$ of the sub-network~$l_{1:i+1}$. This mechanism is described more in detail in Figure~\ref{fig:encoding} b).

\subsubsection{Hyperband++ Engine: }
\paragraph{The original Hyperband algorithm:}
Hyperband \cite{Li17} proposed by Li et al., speeds up random search by using early stopping strategy to allocate resources adaptively. It is easy to use and of good performance, lots of work have based on it. \cite{Falkner18} replaces the random selection of configurations at the beginning of each Hyperband iteration by using Tree Parzen Estimator (TPE) \cite{Bergstra11}. In order to adequately explore large hyper-parameters spaces, \cite{li2018massively} considers the massive parallel hyper-parameters search, and scales linearly with the number of workers in distributed settings as well as converges to a high quality configuration.
However, all these methods focus on hyper-parameters tuning. In our work, we extend the Hyperband to support joint neural network search and hyper-parameters search.

\begin{figure}[!t]
\includegraphics[width=0.45\textwidth]{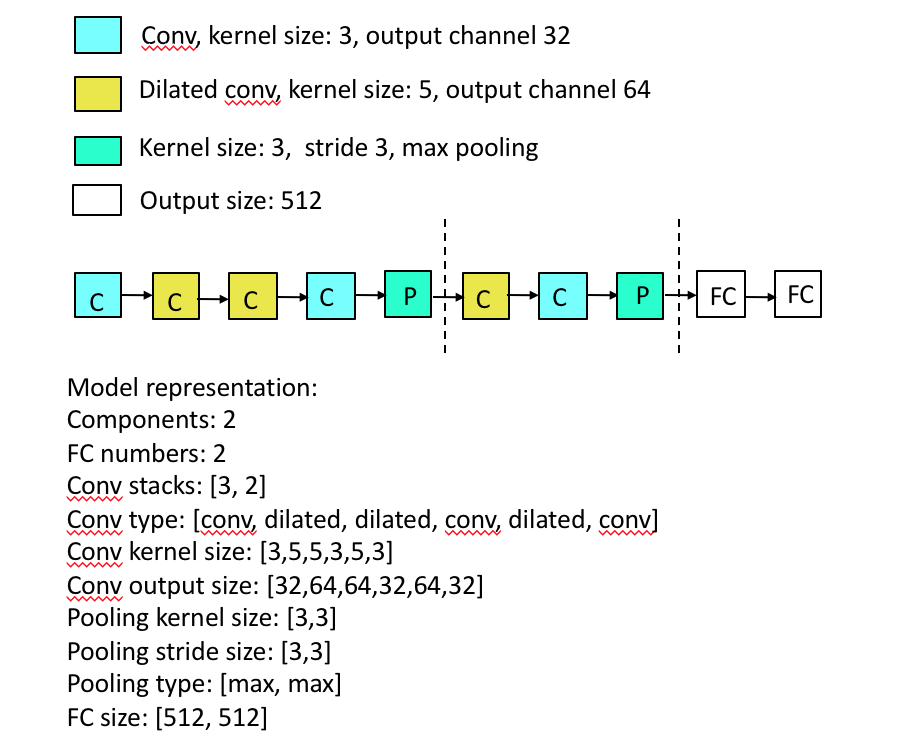}
\caption{Example of chain structure representation supported by Hyperband++}
\label{fig:chain_repr}
\end{figure}

\begin{figure}[!t]
\includegraphics[width=0.45\textwidth]{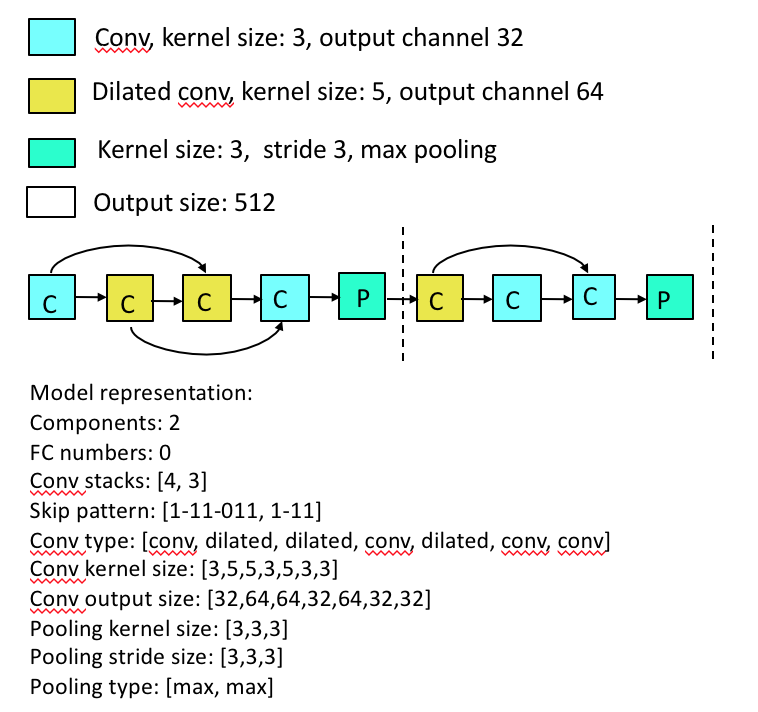}
\caption{Example of skip structure representation supported by Hyperband++}
\label{fig:skip_repr}
\end{figure}

\begin{figure}[!t]
\includegraphics[width=0.45\textwidth]{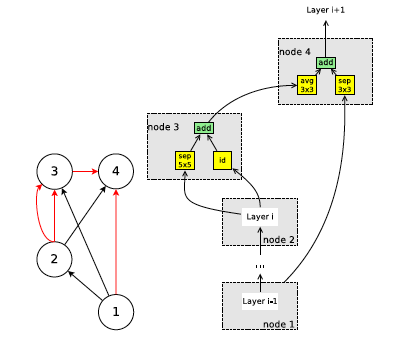}
\caption{Example of multi-branches based structure representation \cite{Pham2018} supported by Hyperband++}
\label{fig:cell_repr}
\end{figure}

\begin{figure}[!t]
\includegraphics[width=0.45\textwidth]{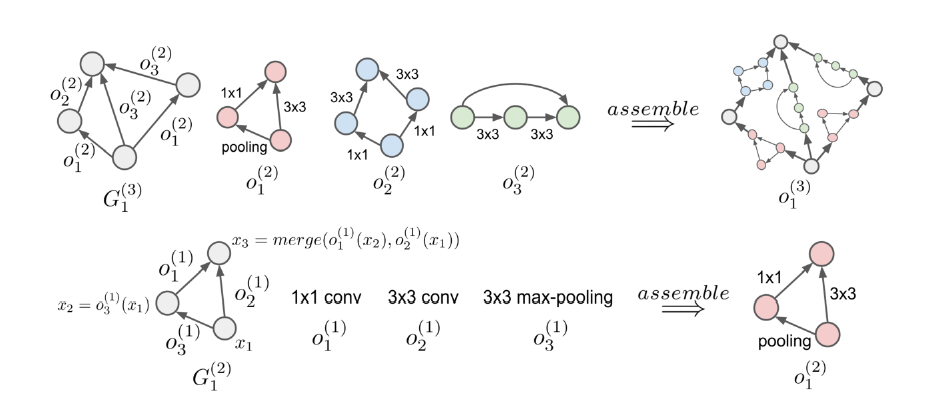}
\caption{Example of hierarchy structure representation \cite{Liu18b} supported by Hyperband++}
\label{fig:hier_repr}
\end{figure}

\paragraph{Model Representation: } 
Effective model representation is necessary to link Hyperband and NAS (Neural Architecture Search). \cite{Liu18b} shows that with an effective model representation, even random search can achieve good performance. In our work, we support four model representations: plain chain structure, skip chain structure, multi-branch structure and hierarchy structure. \textbf{Plain chain structure} is shown in Fig. \ref{fig:chain_repr}. The architecture includes one or more than one components sequentially connected, each component ends with a pooling layer. While if there is no pooling layer in the architecture, it is deemed as of one component. In every component, there will be several convoluational stacks, with attributes of kernel size, type, and output channel numbers. Using this representation, chain structure network and Hyperband search space can be an one to one mapping. As shown in Fig.\ref{fig:skip_repr}, \textbf{Skip chain structure} is similar to plane chain structure where only skip pattern is added. Here we make a constrain that skipping only occur within the component. \textbf{Multi-branches structure} in Fig. \ref{fig:cell_repr} is introduced in \cite{Pham2018}. And it is widely used in the neural network search strategy \cite{Liu18} for cell based search. \textbf{Hierarchy structure} in Fig.\ref{fig:hier_repr} is proposed by \cite{Liu18b}, which defines the three-level hierarchical architecture. See the bottom row, the level-1 primitive operations like convoluational layer, pooling layer etc. are assembled to form a level-2 motif; again various level-2 motifs are assembled to form a level-3 motif as shown in top row in figure. Our work extend the Hyperband to support these four kinds of structures to do the neural network search, we call it Hypterband++. Because Hyperband also have the intrinsic capability to search learning related hyper-parameters like learning rate, weight decay, momentum, our proposed Hyperband++ can do joint search for both neural network and hyper-parameters.

\paragraph{Meta-learning for Hyperband: }
\label{sec:mlfh}
As we known, most existing Neural network approaches take considerable long time for model searching. Thus there are lots of methods target high efficient NAS, such as weight sharing \cite{Pham2018}, one-shot model \cite{Liu18b}, multi-task Bayesian \cite{Swersky13} and performance prediction\cite{Baker17b}. In our work, we propose a method to reuse the neural network or part of the neural network to speed up the searching process. The method is tailored to the setting whereby the datasets come sequentially and one need to search model for the new arrival datasets efficiently. First, the meta-features of datasets can be extracting, then we implement the virtual dataset layer and group the datasets based on the meta-features at the first step. Accordingly, we follow the general idea of recycle and reuse, such that the pre-searched models i.e. architectures and hyper-parameters can be used for the new arrival datasets. It should be noted that the meta-learning technique can also be coupled with most NAS methods (not limited on Hyperband) in an out-of-box fashion.

\paragraph{Hyperband for Object Detection: } 
Unlike image classification, object detection also considers informative region selection and feature extraction besides classification. Traditional object detection methods are built on handcrafted features and shallow trainable architectures. However, their performance is limited as low-level features can not represent the high-level context from object detectors effectively. Thanks to the rapid development in deep learning, semantic, high-level features can be learned. There are mainly two types of frameworks in generic object detection: region proposal based methods, which include RCNN \cite{girshick2014rich}, FRCNN \cite{girshick2015fast}, Faster R-CNN \cite{ren2015faster}, FPN \cite{lin2017feature} and Mask R-CNN \cite{he2017mask}, etc.; Regression/Classification based methods, which contain YOLO \cite{redmon2016you}, SSD \cite{liu2016ssd}, YOLOv2 \cite{redmon2017yolo9000}, etc. Empirical results show the performance of object detection for different datasets is highly depended on the pre-trained network and lots of hyper-parameters. In this section, we describe the ways to apply Hyperband for object detection.

\textbf{Hyper-parameters tuning for Object Detection: } Comparing with image classification, there are more hyper-parameters in object detection applications. Besides learning process related hyper-parameters like learning rate, decay policy, momentum, etc., there are also object detection related hyper-parameters including anchor size, aspect, pre-trained model categories, loss weights, fraction of foreground, overlap for positive RPN, bbox threshold, FPN level, etc.. Hyperband can be used to deal with tuning these kinds of hyper-parameters efficiently.

\textbf{Meta-learning for high efficient Hyper-parameters tuning: } By using the methods described in \ref{sec:mlfh}, hyper-parameters tuning for object detection can be speeded up by grouping the similar datasets with the same set or subset of hyper-parameters. In object detection, there are also lots of works by adopting the concept of groups. \cite{lenc2015r} uses a biasing sampling to match the statistics of the ground truth bonding boxes with K-means clustering, while \cite{xiang2017subcategory} proposes a subcategory-aware RPN. We leave more effective groups construction methods beyond hyper-parameter sharing for future work.

\textbf{Architecture Search for Object Detection: } Considering the object detection pipeline, with newly-searched network architecture, it always needs pre-train on ImageNet to obtain the corresponding reward values, which makes it impossible to search neural network for object detection. An alternative way is to use transferable network to replace the existing pre-trained networks like Inceptions, resNet, VGG and ZF. \cite{Liu18b} shows that the auto-searched network for CIFAR10 can be used as pre-trained network of Faster-RCNN for object detection. In our work, we first use Hyperband to search the optimal architectures for different datasets for image classfication task, and then plug in these architectures pre-trained on ImageNet into Faster-RCNN or FPN pipeline. By using the meta-learning proposed in section \ref{sec:mlfh}, the proper pre-trained network are chosen based on the input dataset.

\subsection{Fine grained synthesis}
\label{sec:fine-grained-algo}
\vspace{1cm}
Broadly, we describe a supervised learning algorithm that serves as an optimization step towards automatically generating a neural network model. Specifically, given an input dataset with classification labels, we will describe an algorithm to incrementally add new connections and a small number of new trainable variables within a neural network to improve the prediction accuracy of the neural network.

There are a lot of techniques that use template layers such as convolution, fully-connected, max-pool, avg. pool, etc to automatically synthesize neural networks. These techniques work well for pre-processed and cleaned datasets but due to their parameter size have a tendency to over-fit to the training data.
There is another set of techniques that explores more "fine-grained" connections within neural networks. This class of techniques are more along the lines of the proposed technique. The technique proposed in [Filter-Shape] computes a co-variance metric to determine which features (inputs) that should be combined to create a filter.

The filter shaping approach is restricted by the computational complexity of computing co-variance matrix. Hence the size of the neighborhood that can be searched using this technique is smaller than the proposed technique. Secondly, the filter shaping paper looks for features with the most correlation. However, it is unclear whether such a metric is fundamentally necessary for good generalization.

Our technique uses an evolutionary algorithm for building a custom convolution filter. The algorithm has five phases -
0) Given a network that has been trained until a certain early stopping criteria is met, the 0th or initialization phase involves selecting a subset of neurons that are "important" for the given dataset and classification task . If there is no network to start with then a single fully connected layer connecting the inputs to the outputs is used as a starting network.
1) The first phase involves growing the network by adding a layer of dense connections to this subset of neurons. The initial values of the new connections are chosen such that the output neuron values are not perturbed. Doing so ensures that the new connections do not cause the network to forget what has been learnt. The newly grown network is then re-trained for a few epochs.
2) The second phase involves iteratively pruning network connections in the newly grown layer that show the least change from their initialized values. During this pruning stage the network is trained for a few epochs after every pruning step.
3) The third phase involves merging connections in the newly grown layer into "k" weight buckets. The degree of similarity of input weight distribution determines whether the connections are merged into a single bucket or not. The "k" buckets then become the "k" custom filters of the newly grown layer.
4) The fourth and final phase involves re-initializing the weights of the pruned and merged network so that the output neuron values are unperturbed from the values after phase (i). The new network is then retrained until the early stopping criteria is met.

The proposed algorithm learns the shape of the custom filters from the evolution of the weight values on the connections. Due to this it is able to leverage the acceleration in training offered by specialized hardware such as GPUs. Thus, in comparison with techniques such as [Filter-Shape] and [Grow-Prune] the proposed approach is much faster. Another advantage of this technique over [Filter-Shape] is that it offers higher accuracy improvement with the same increase in network size, which translates to overall better network quality.

The network shown in Phase 0 represents either:
a) The final fully connected layer of an existing network that is either manually designed or auto-generated from another neural architecture search algorithm.
or
b) As a more general application of the proposed technique, the blue input neurons can represent a subset of "important" neurons from an existing neural network. If an initial network is not available, the blue input neurons can also represent all or a subset of input features of the given dataset. In this case, a network is initialized with a fully connected layer connecting blue input neurons to the red output neurons.

Assuming that there are 'n' input neurons and 'm' output neurons, the operation performed in the fully connected layer can be expressed as:

y = Act(W*x + B)

where 'x' is the 'n'-dimensional vector containing values of the input neurons, y is the 'm'-dimensional vector containing values of the output neurons, W is the 'mxn'-dimensional matrix containing weight variables, B is the 'm'-dimensional vector containing bias variables, and Act is an activation function for e.g., Relu, Tanh, Sigmoid, etc.

The initial values of W and B are obtained by training the initial network until the early stopping criteria is met.

In phase 1, the network is grown by adding a hidden layer containing 'l' hidden neurons. The layer connecting input neurons to the hidden neurons can either be fully connected or selectively connected to combine input neurons selectively into a hidden neuron. The operation performed by this layer can be expressed as:

z = Act(W'*x + B')

where z is an 'l'-dimensional vector containing values of the hidden neurons, W' is an 'lxn'-dimensional weight matrix, B' is an 'l'-dimensional bias vector, Act is the activation function. In the case where the layer is selectively connected, the missing connections can be represented as zeros in the weight matrix.

The layer connecting hidden neurons to the output neurons is a fully connected layer. The operation performed by this layer can be expressed as:

y = Act(W''*z + B'')

where y is an 'm'-dimensional vector containing values of the output neurons, W'' is an 'mxl'-dimensional weight matrix, B'' is an 'm'-dimensional bias vector, Act is the activation function.
The initialization of the two layers in phase 1 has to be carefully determined to ensure continuity in training an evolving network in phase 0. To achieve this, the initial values for [W',B'] and [W'', B''] are derived from the trained values of [W, B] at the end of phase 0. The initial values satisfying this constraint can be achieved in many different ways, for e.g., assume a blue input layer neuron gets connected to 3 neurons in the hidden layer. The weight variable for these neurons can be
initialized to w1, w2, w3 such that sum(w1, w2, w3) = 1. The weight variable for connections from other neurons in the input layer to these 3 neurons can be initialized to 0. Such an initial value assignment ensures continuity in the training process.

In phase 2 of the optimization, connections in the layer between the input and hidden layer are pruned based on a certain pruning metric. Examples of metrics for pruning include value of weight variable, absolute value of weight variable, magnitude of change in weight value over several epochs. Pruning can happen either as a single step or in multiple steps applied iteratively. Between two pruning steps, the network is trained for a few epochs for the values to adjust for the pruned variables. At the end of pruning, N input connections are retained for each hidden neuron.

In phase 3 of the optimization the N input weight variables for the neurons in the hidden layer are merged into 'k' buckets of N weight variables each. Merging of weight variables takes place based on the similarity of shape of the distribution of the weight variables. One way of measuring similarity is to use the L-2 distance of the normalized values of weight variables. For instance, [1.2, 0.6, 0.3] has a shape similar to [2, 1, 0.5] than to [1, 0.9, 0.8].

Finally, in phase 4, the merged and pruned network weights and bias values are re-initialized such that the values of output neurons are unperturbed from phase 0. The re-initialized network is then trained until the early stopping criteria is met.


\section{Experimental Evaluation}\label{sec:experiments}

We tested the NeuNetS Framework on various benchmark datasets for image and text classification.


All images are normalized by subtracting the mean and dividing by the standard deviation.
For images with resolution higher than 64x64, they are scaled to 64x64 for NCEvolve.
For TAPAS, images are always scaled to 32x32.
The maximum GPU budget per dataset is divided into three categories: low, medium and high.
A dataset is assigned to one category based on the number of examples.
All datasets with at most 10K examples get a low GPU budget of 2 hours.
Datasets with at most 75K examples get the medium budget of 5 hours.
Finally, all other datasets get the high budget of at most 16 hours.
In contrast to the literature in the domain of automated architecture search, this budget contains both the search and training time.
State-of-the-art methods use double this budget only for the search followed by an expensive post-processing \cite{Pham2018}.
We evaluate NeuNetS on 12 image classification benchmarks and report the results in Table \ref{tab:results-images}.

As we all know, machine learning models cannot accept text as input. They only work with integers or floats. In order to overcome this, a standard practice is to
tokenize the training data and identify the most frequent $K$ words (excluding stop words) and  map them to integers. To elaborate further,
the most common word would be given an integer representation of 0, the second most common word a representation of 1, and so on. All the words outside of the
top $K$ are replaced by an unknown token \textit{UNK}. The model requires input of fixed dimensions. thus, the maximum number of tokens \textit{MAX} in the input in predetermined.
If the number of words in an instance is greater than the maximum imposed, the instance is truncated to \textit{MAX} length. If the number of words is less than \textit{MAX}, we pad the sentence with \textit{UNK} tokens to meet the desired length.
The weights of the first layer of the deep learning model, also known as the embedding layer, are initialized with the word embedding matrix of the top K words in the training data.
The $i^{th}$ row of the embedding matrix represents the word embedding (obtained from GloVe~\cite{pennington2014glove} or Word2vec~\cite{mikolov2013efficient, mikolov2013distributed, mikolov2013linguistic}) of the word whose integer mapping is $i$. 

Based on similar criteria, we also impose maximum GPU budget for the synthesis of text classifiers.
Therefore, we assign 2 hours for a dataset comprising of at most 250K examples, 5 hours for those up to 2M examples and a maximum of 16 hours for the rest. We evaluate NeuNets on 9 text classification datasets and report the results in Table~\ref{tab:results-text}.

\begin{table}[t]
\caption{Results on various image classification benchmarks with required training time in GPU hours.\label{tab:results-images}}
\centering
\begin{tabular}{lrrrrr}
\hline\noalign{\smallskip}
Dataset & Cls & Examples & Error & Time & Params\\
\noalign{\smallskip}
\hline
\noalign{\smallskip}
Caltech-256 \cite{Griffin2007} & 257 & 31K & 48.56 & 5.0 & 5.78M\\
CIFAR-10 \cite{Krizhevsky2009} & 10 & 60K & 6.32 & 3.6 & 3.68M\\
CIFAR-100 \cite{Krizhevsky2009} & 100 & 60K & 27.79 & 3.9 & 9.60M\\
Fashion \cite{Xiao2017} & 10 & 70K & 4.51 & 3.2 & 5.73M\\
Flowers-5 \cite{Nilsback2008} & 5 & 4K & 16.41 & 2.0 & 4.57M\\
Flowers-102 \cite{Nilsback2008} & 102 & 2K & 54.02 & 1.2 & 3.35M\\
Food-101 \cite{Bossard2014} & 101 & 101K & 38.12 & 11.9 & 7.38M\\
GTSRB \cite{Stallkamp2011} & 43 & 52K & 3.52 & 4.1 & 3.05M\\
MNIST \cite{Lecun1998} & 10 & 70K & 0.64 & 2.5 & 3.74M\\
Quick, Draw! \cite{Ha2017} & 345 & 380K & 27.34 & 16.0 & 2.58M\\
STL-10 \cite{Coates2011} & 10 & 13K & 25.14 & 1.8 & 6.68M\\
SVHN \cite{Netzer2011} & 10 & 99K & 3.37 & 12.6 & 4.83M\\
\hline
\end{tabular}
\end{table}


\begin{table}[t]
\caption{Results on various text classification benchmarks with required training time in GPU hours.\label{tab:results-text}}
\centering
\begin{tabular}{lrrrrr}
\hline\noalign{\smallskip}
Dataset & Cls & Examples & Error & Time\\
\noalign{\smallskip}
\hline
\noalign{\smallskip}
Cola~\cite{warstadt2018neural} & 2 & 9K & 29.60 & 1.1\\
IMDB Sentiment~\cite{maas2011learning} & 2 & 22K & 12.00 & 0.7\\
Rotten TMC~\cite{rotten} & 5 & 140K & 31.51 & 0.9\\
SMS Spam~\cite{almeida2013towards} & 2 & 5K & 0.54 & 0.2\\
Snips~\cite{coucke2018snips} & 7 & 2K & 0.00 & 0.2\\
Stanford Sentiment~\cite{SocherEtAl2013} & 6 & 215K & 31.17 & 1.0\\
TREC~\cite{xin2002learning} & 6 & 5K & 11.62 & 0.3\\
Yelp~\cite{yelp} & 5 & 52K & 40.45 & 1.0\\
Youtube Spam~\cite{alberto2015tubespam} & 2 & 2K & 3.05 & 0.2\\
\hline
\end{tabular}
\end{table}



\bibliographystyle{IEEEtran}
\bibliography{neunets}

\end{document}